\documentclass[conference]{IEEEtran}
\IEEEoverridecommandlockouts
\usepackage{cite}
\usepackage{amsmath,amssymb,amsfonts}
\usepackage{algorithmic}
\usepackage{graphicx}
\usepackage{textcomp}
\usepackage{xcolor}
\usepackage{booktabs} 
\usepackage{multirow} 
\usepackage{url}
\usepackage{hyperref}
\usepackage{makecell}

\def\BibTeX{{\rm B\kern-.05em{\sc i\kern-.025em b}\kern-.08em
    T\kern-.1667em\lower.7ex\hbox{E}\kern-.125emX}}
\begin{document}

\title{Underrepresented in Foundation Model Pretraining Data? A One-Shot Probe
}
\author{\IEEEauthorblockN{
Chris Vorster, 
Mayug Maniparambil, 
Noel E. O'Connor,
Noel Murphy, 
Derek Molloy \\
\textit{ML-Labs, Dublin City University, Dublin, Ireland} \\
}
}

\maketitle

\begin{abstract}
Large-scale Vision-Language Foundation Models (VLFMs), such as CLIP \cite{b7}, now underpin a wide range of computer vision research and applications. VLFMs often are adapted to various domain-specific tasks \cite{MediCLIPAdaptingCLIP2024_zhang, WildCLIPSceneAnimal2024_gabeff, RemoteCLIPVisionLanguage2024_liu}. However, VLFM performance on novel, specialised, or underrepresented domains remains inconsistent. Evaluating VLFMs typically requires labelled test sets, which are often unavailable for niche domains of interest, particularly those from the Global South. We address this gap by proposing a highly data-efficient method to predict a VLFM's zero-shot accuracy on a target domain using only a single labelled image per class. Our approach uses a Large Language Model to generate plausible counterfactual descriptions of a given image. By measuring the VLFM's ability to distinguish the correct description from these hard negatives, we engineer features that capture the VLFM's discriminative power in its shared embedding space. A linear regressor trained on these similarity scores estimates the VLFM's zero-shot test accuracy across various visual domains with a Pearson-r correlation of 0.96. We demonstrate our method's performance across five diverse datasets, including standard benchmark datasets and underrepresented datasets from Africa. Our work provides a low-cost, reliable tool for probing VLFMs, enabling researchers and practitioners to make informed decisions about data annotation efforts before committing significant resources. The model training code, generated captions and counterfactuals are released \href{https://github.com/chris-vorster/PreLabellingProbe}{here}.
\end{abstract}


\section{Introduction}
The \textit{Machine Learning} field is undergoing a paradigm shift, moving away from task-specific models towards more universal solutions known as Foundation Models \cite{b1}. These large-scale models, such as BERT \cite{devlin2019bert}, DALL-E \cite{ramesh2021dalle}, CLIP \cite{b7}, and the family of GPT models \cite{brown2020gpt3, openai2023gpt4}, are trained on vast, diverse datasets, often self-supervised, and demonstrate a remarkable capacity to generalise across a wide array of downstream tasks. In \textit{Computer Vision}, this trend challenges the long-standing ImageNet-based methodology \cite{b2, b3}, where models are first pretrained on a large, supervised dataset and then fine-tuned for specific applications \cite{b4, b5, b6}.

Vision-Language Foundation Models (VLFMs) like CLIP \cite{b7} and ALIGN \cite{b8} have come to define this new era. By learning a shared embedding space from hundreds of millions or even billions of image-text pairs scraped from the web, these models have become the \textit{de facto} standard for \textit{zero-shot} image recognition and image-text retrieval \cite{b9, b10}. Their power lies in leveraging natural-language prompts (e.g., ``A photo of a \{label\}") to create classifiers dynamically, eliminating the need for task-specific training data \cite{b11}.

However, the zero-shot capability of VLFMs is tied to the distribution of concepts in their massive, yet noisy and coarse-grained, training data \cite{b12}. Analysis reveals that concept frequency in these web-scale datasets follows a \textit{Zipfian}, long-tailed distribution \cite{b13, b9}. Consequently, models exhibit a log-linear performance scaling: they require exponentially more data on a concept to achieve linear gains in performance \cite{b9}. This means a vast number of concepts remain underrepresented, leading to poor performance on tasks requiring fine-grained distinctions or knowledge of concepts outside the mainstream internet context.

This data-centric limitation has important implications for equity and inclusion in \textit{AI}. A significant barrier to AI adoption in regions like Africa is the scarcity of datasets relevant to local challenges in agriculture, healthcare, and culture \cite{b14}. When Foundation Models, predominantly trained on data from the Global North, are applied to these contexts, they often fail. This perpetuates a cycle of ``data colonialism" \cite{b14}, where control over the data and algorithms that shape AI systems remains concentrated, and solutions are not tailored to diverse global needs. While initiatives exist to diversify datasets, they often fail to capture the deep cultural nuances necessary for robust performance \cite{b14}.

The core problem is foresight: how can we know if a Foundation Model will perform well on a specific domain without collecting a large, (likely) expensive, and time-consuming test dataset? 

\begin{figure*}[t]
\centerline{\includegraphics[width=\textwidth]{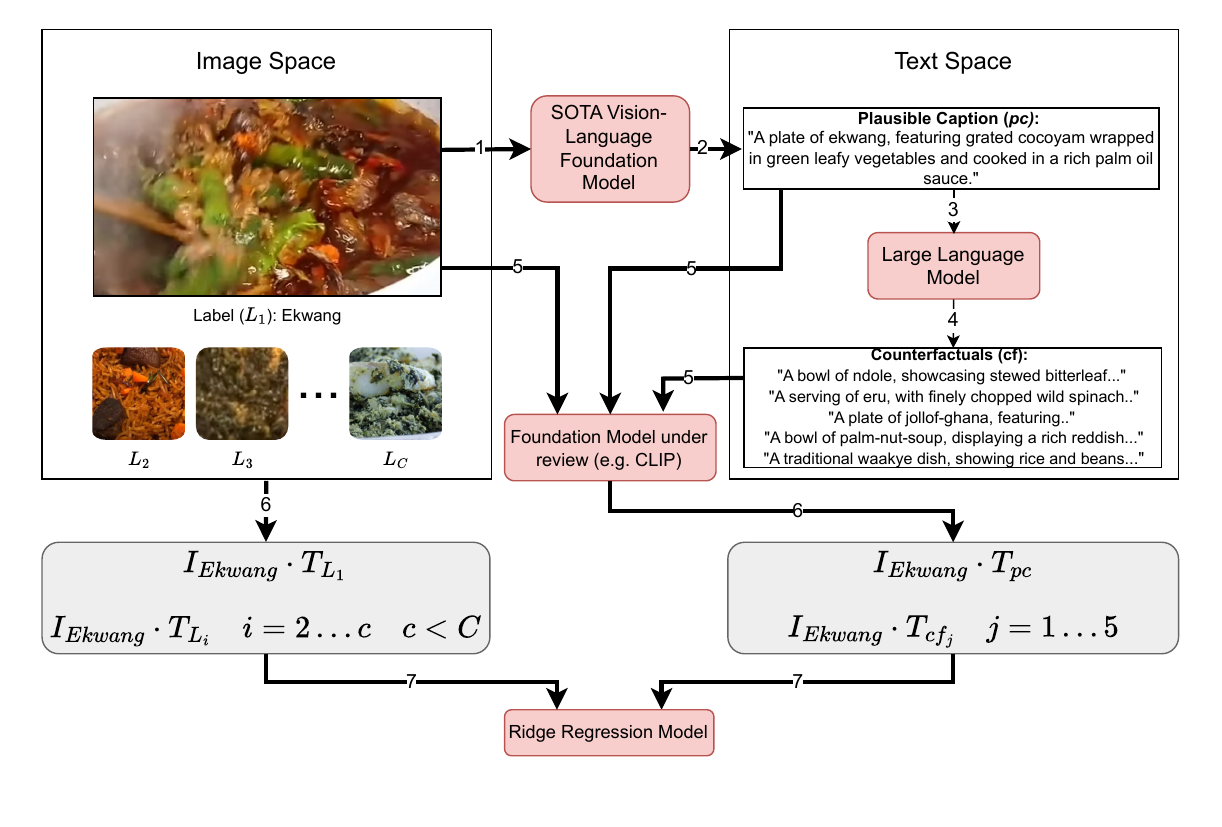}}
\caption{Methodological Overview. This figure illustrates our three-stage pipeline for predicting a Vision-Language Model's zero-shot accuracy on a target domain, using the ``Ekwang" class from the African Food dataset as an example. Counterfactual Probing: A single representative image (Step 1) is used to generate a plausible caption $T_{pc}$ (Step 2) and a set of counterfactual captions $\{T_{cf_i}\}$ via an LLM (Steps 3-4). Similarity Scoring: The VLFM under evaluation is used to compute embeddings for the image ($\mathcal{I}$) and captions (Step 5). $\{T_{L_i}\}$ represents the standard CLIP zero-shot text prompt for class $i$, e.g. ``a photo of Ekwang". Two sets of similarity scores are calculated: one for the standard CLIP zero-shot prompts and another for the LLM-generated captions (Step 6). Performance Prediction: The similarity scores are used as input to a Ridge Regression Model. (Steps 7). The model is trained to estimate the VLFM's zero-shot accuracy on the full test set using only one labelled image per class.}
\label{fig:method_overview}
\end{figure*}

To address this challenge, we introduce a novel, data-efficient method for predicting a VLFM's zero-shot transfer performance. Our key insight is that a model's global performance on a dataset can be inferred by testing its local understanding of individual concepts. Using a single image per class, we use a Large Language Model (LLM) to generate a plausible caption and a set of challenging, semantically related, but incorrect \textbf{counterfactual prompts} - see Fig. \ref{fig:method_overview} for an example. By evaluating the VLFM's ability to distinguish the true description from these distractors, we probe the geometry of its embedding space. We use the VLFM's similarity scores and a linear model to predict the dataset-level test set zero-shot accuracy. Our approach provides a low-cost, reliable estimate of a VLFM's capabilities, empowering users to evaluate models for their specific needs, especially in under-resourced or specialised domains.

\section{Related Work}
\subsection{Performance Prediction and Out-of-Distribution Detection}
Our work is conceptually related to Out-of-Distribution (OoD) detection, which aims to identify inputs that differ from the training distribution. Robust OoD detection is essential in safety-critical domains such as autonomous driving and healthcare, where misclassifying unknown samples can have severe consequences \cite{b15}. Ding \textit{et al}. \cite{b16} use a large set of auxiliary ``outlier" class labels to create pseudo-OoD prompts, refining them into prototypes to score OoD likelihood. The ZOC method \cite{b15} extends CLIP \cite{b7} with a text generator to create candidate unseen labels, performing OoD detection without prior knowledge of unseen classes. Related work by Saxena \textit{et al}. \cite{b17} has shown a strong linear correlation between a model's in-distribution (ID) and OoD accuracy. This enables OoD performance prediction using only unlabelled data by measuring model agreement.

\textbf{Our Distinction:} While these methods focus on detecting individual OoD samples or predicting performance based on ID-OoD correlations, our goal differs. We aim to predict the VLFM's zero-shot classification accuracy without access to a full test set of image-text pairs for the domain of interest. Our approach does not require ID data or a set of outlier labels; instead, we generate targeted, semantic ``hard negatives" (counterfactuals) per class to directly probe the model's discriminative power.

\subsection{Counterfactual Reasoning in Vision-Language Models}
The use of counterfactuals or `foils' to test model understanding has been explored in vision-language research. Shekhar \textit{et al.} \cite{b15} introduced the FOIL-COCO dataset in which human-annotated captions are minimally altered by replacing a single word, creating a subtle mismatch with the image. This dataset challenges models to perform fine-grained error detection.

\textbf{Our Distinction:} Similarly, we leverage the idea of generating known errors to probe a model. However, our method differs in both mechanism and goal. We use the generative power of LLMs to create a diverse set of semantically plausible but incorrect descriptions, whereas Shekhar \textit{et al}. \cite{b15} used a Long Short-Term Memory model. More importantly, our goal is not to detect the mismatch but to use the VLFM's similarity scores between the image, a plausible caption, and its counterfactuals as signals to predict the VLFM's overall zero-shot performance on the dataset.

\subsection{Underrepresentation in Vision-Language Models}
Recent research has quantified the impact of data distribution on VLFM performance. Udandarao \textit{et al}. \cite{b9} demonstrated that multimodal models are sample-inefficient, requiring an exponential increase in concept frequency during pre-training to achieve linear gains in downstream performance. Their analysis confirmed that concept distributions are extremely long-tailed, reminiscent of Zipf's law \cite{b13}. The frequency of a concept in the pretraining set thus serves as a powerful predictor of model performance on that concept. However, for proprietary VLFMs, we do not have access to the underlying pretraining datasets, which makes it difficult to verify these distributions. Even when open-source datasets are used, measuring concept frequency is not straightforward, since concepts may be expressed in many ways (e.g., polysemy, synonymy, or noise in web-crawled alt-text).

\textbf{Our Distinction:} Our counterfactual method serves as a probe for the effects of the hidden pretraining frequency. By assessing how well-defined the representation is for a given concept, we indirectly measure the quality of the model's training data on that concept, allowing us to predict its downstream performance without needing access to the pretraining dataset.

\section{Methodology}
Our method aims to determine how distinctive a given VLFM's representation is for a specific domain (i.e., a set of classes). We achieve this with high data efficiency by requiring only one labelled image per class. The methodology is structured into three stages as shown in Fig. \ref{fig:method_overview}.

\subsection{Counterfactual Probing of the Shared Embedding Space}
We probe the local geometry of the VLFM's shared image-text embedding space as follows:

\textbf{Image-to-Language Anchoring.} We randomly select one image per class from a diverse set of commonly used datasets. We use a multimodal model (e.g., GPT-5-Nano \cite{ModelOpenAIAPI_}) conditioned on the image and its ground-truth label to generate a high-quality, plausible caption $T_{pc}$ aligned with the image's content.

\textbf{Generating Counterfactuals.} With $T_{pc}$ as an anchor, we use a (text-based) LLM to generate $N$ counterfactual captions, $T_{cf_{i}}$. The LLM is prompted to generate descriptions that are semantically related to $T_{pc}$ but correspond to other visually confusable classes within the same dataset. This creates a set of ``hard negatives." For our experiments, we set $N=5$.

\textbf{VLFM-based Similarity Scoring.} Using the VLFM to be evaluated (e.g., OpenCLIP \cite{ReproducibleScalingLaws2023_cherti}), we compute normalized embeddings for the image (${I_j}$), plausible caption ($T_{pc}$), and counterfactual captions ($T_{cf_i}$). We then measure the cosine similarity, yielding a similarity score, $s_{pc_j} = {I_j} \cdot T_{pc}$, and a list of counterfactual scores, $s_{cf_{ij}} = {I_j} \cdot T_{cf_i}$.

\subsection{Vanilla Zero-Shot}
We use the standard zero-shot similarity scores created using the prompt: ``A photo of a \{label\}", yielding $c$ text embeddings $T_{L_{i}}$. In our experiments, we set $ c = N$, i.e., we randomly select the same number of label-based prompts as counterfactuals. For each query image (one per class), we thus have twelve similarity score values as features for the Ridge Regression Model. Method design choices are ablated in Section~\ref{sec:design_options}.

\subsection{Transfer Performance Prediction}
The final stage uses the twelve similarity-based features to estimate the VLFM's Zero-Shot accuracy. We train a linear regression model on a diverse collection of datasets, where ground-truth targets are the VLFM’s measured Zero-Shot accuracies on full test sets. We use Ridge Regression because it constrains coefficient magnitudes via L2-regularisation to address feature correlation, as is the case in our setting. Once trained, the model can be applied to a new domain using only one labelled image per class, providing a data-efficient estimate of transfer performance.
\begin{table*}[t]
\centering
\caption{Overview of Datasets Used}
\label{tab:datasets}
\resizebox{\textwidth}{!}{%
\begin{tabular}{@{}l|l|l|p{9cm}@{}}
\toprule
\textbf{Dataset} & \textbf{Task} & \textbf{Classes} & \textbf{Focus} \\
\midrule
CIFAR-10 \cite{krizhevsky2009learning} & Object classification & 10 & Ten common object categories (airplane, automobile, bird, etc.) in low-resolution natural scenes. \\
CIFAR-100 \cite{krizhevsky2009learning} & Object classification & 100 & One hundred fine-grained classes grouped into 20 super-classes (e.g., maples, lobsters). \\
Oxford 102 Flowers \cite{nilsback2008automated} & Flower classification & 102 & Photographs of 102 flower species with diverse backgrounds, lighting, and viewpoints. \\
Food-101 \cite{bossard2014food} & Food classification & 101 & Real-world dishes (e.g., pizza, sushi, ramen) with varied preparation and noisy backgrounds. \\
Caltech-101 \cite{fei2004learning} & Object classification & 101 & Images of objects from 101 categories (e.g., faces, motorcycles, crabs) with varying scale, position, and background clutter. \\
Oxford-IIIT Pet \cite{parkhi2012cats} & Pet breed classification & 37 & Cat and dog breeds with diverse poses, backgrounds, and bounding box annotations. \\
DTD \cite{cimpoi2014describing} & Texture classification & 47 & Texture attributes (e.g., striped, dotted, wrinkled) from various materials and conditions. \\
EuroSAT \cite{helber2019eurosat} & Land-use classification & 10 & Sentinel-2 patches of land-use categories (e.g., residential, forest, river) across Europe. \\
GTSRB \cite{stallkamp2012man} & Traffic-sign class. & 43 & German road signs in real driving scenarios with weather, occlusion, and angle changes. \\
FGVC Aircraft \cite{maji2013fine} & Aircraft-model class. & 100 & Commercial and military aircraft variants distinguished by subtle features. Regarded as a fine-grained image classification dataset. \\
ImageNet-v2 \cite{recht2019imagenet} & Distribution-shift test & 998 & Re-collected version of ImageNet-1k to measure generalisation under data distribution shifts. \\
MNIST \cite{lecun1998gradient} & Handwritten digit class. & 10 & Classic benchmark for handwritten digit recognition, consisting of centred, low-res digits 0–9. \\
STL-10 \cite{coates2011analysis} & Object classification & 10 & A benchmark for semi-supervised learning with higher-res images and a large unlabeled set. \\
ImageNet-O \cite{hendrycks2021natural} & Robustness/OoD evaluation & 200 & Images from 200 object categories not included in ImageNet-1k, used to test out-of-distribution handling. \\
African Food \cite{omeiza2022african} & Food classification & 6 & Traditional African dishes (e.g., jollof rice, injera), highlighting cultural representation in food datasets. \\
Beans \cite{makerere2021beans} & Plant disease class. & 3 & Images of healthy and diseased bean leaves for agricultural disease diagnosis in smallholder farms. \\
\bottomrule
\end{tabular}%
}
\end{table*}
\section{Experiments and Results}

\subsection{Experimental Setup}
\textbf{Datasets.} We evaluate our method on a diverse suite of datasets (see Table \ref{tab:datasets}), covering generic and fine-grained classification, specialised domains, and African datasets. Specifically, we include two African-focused datasets \textbf{African Food} and \textbf{Beans} to test performance on underrepresented domains.

\textbf{Implementation Details.} We evaluate the OpenCLIP-ViT-B/16 model pretrained on LAION-400M\cite{schuhmann2021laion400m}. We use the GPT-5-Nano model \cite{ModelOpenAIAPI_} for caption and counterfactual generation. We set $N=5$ (number of counterfactuals per class). Our linear regression model is trained on eleven datasets (blue points in Fig. \ref{fig:results_scatter}) and tested on five hold-out datasets.

\subsection{Prediction Accuracy}
\begin{figure}[t]
\centerline{\includegraphics[width=\columnwidth]{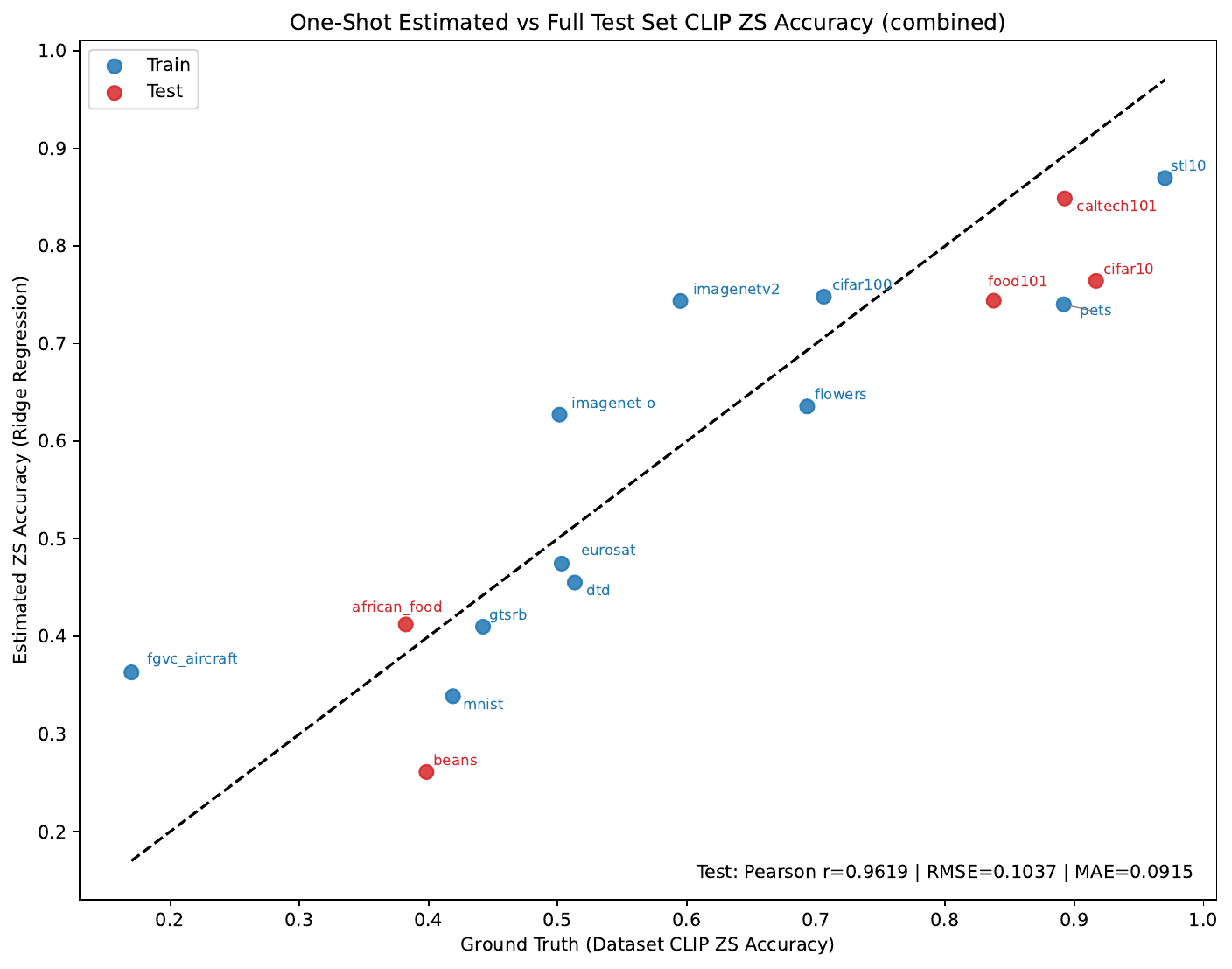}}
\caption{Ground-Truth vs. Predicted Zero-Shot Accuracy. The scatter plot depicts our method's predictions across 16 datasets. The x-axis shows the actual zero-shot accuracy of OpenCLIP-ViT-B/16, calculated on each dataset's full test set, and the y-axis shows our predicted accuracy using only a single labelled image per class. The dashed black line indicates perfect agreement. Blue points are datasets used to train our Ridge Regression Model; red points are unseen test datasets. The strong test correlation demonstrates our method's accuracy and generalisation, even in underrepresented domains such as \textit{African Food} and \textit{Beans}.}
\label{fig:results_scatter}
\end{figure}

Our central claim is that our one-shot method provides a useful estimate of a VLFM's zero-shot performance. As shown in Fig. \ref{fig:results_scatter}, the predicted and ground-truth zero-shot accuracies exhibit a strong linear correlation across datasets.

The model performs well on both training and, more importantly, hold-out test datasets (red points). The test points lie relatively close to the identity line, indicating robust generalisation to new visual domains. This includes specialised datasets like the African Food \cite{omeiza2022african} and Beans \cite{makerere2021beans} datasets. The strong Pearson-r correlation across the test datasets demonstrates our method's ability to capture the necessary signal in VLFM's similarity scores to estimate dataset-level performance in new domains. As a result, in practice, labelling efforts will not be wasted on datasets that a Foundation Model already comprehends. Furthermore, our approach assists in answering two key questions: (1) Is a specific Foundation Model suitable for a target domain? (2) What is the required annotation granularity?

\section{Ablation and Discussion}
We perform ablations to quantify the contribution of each component in \textit{PreLabellingProbe}. We compare (i) using only LLM-generated captions and counterfactuals, (ii) using only vanilla CLIP prompt-based scores, and (iii) the full combined variant. This isolates the source of gains and tests whether the signals are complementary.

\subsection{Method Design Ablation}
\label{sec:design_options}
To validate the contribution of each component in our method, \textit{PreLabellingProbe}, we conducted an ablation study that isolates each component. Our central hypothesis is that the semantic richness of LLM-generated counterfactuals, along with CLIP's prompt-based scores, provides the necessary signal to probe a VLFM's understanding.

In this ablation, the vanilla prompt-based CLIP \cite{b7} similarity scores are calculated using the prompts ``a photo of \{class\_name\}" and ``a photo of \{other\_class\_name\}", where ``\{other\_class\_name\}" is randomly sampled from the other classes in the dataset. The rest of our pipeline remained unchanged to ensure a fair comparison. 

The results, shown in Table \ref{tab:performance_comparison}, demonstrate that the combined \textit{PreLabellingProbe} variant achieves the strongest agreement with ground-truth accuracies (highest Pearson-r and lowest RMSE), indicating complementary signal between LLM-derived and vanilla prompt features.

\subsection{Resource use}
The cost to evaluate a new dataset using \textit{PreLabellingProbe} is minimal. As an example, for the African Food dataset (6 classes), LLM caption and counterfactual generation takes 1min 23s in total (14s per class) with an API cost of \$0.006. CLIP similarity score computation and Ridge Regression inference take less than 5 seconds on CPU (AMD Ryzen 9 7950X), making the pipeline practical on standard hardware.

\subsection{Results}
We report two complementary evaluation metrics to quantify both relative ranking quality and absolute accuracy.

\noindent\textbf{Pearson-r:} Pearson's correlation coefficient measuring linear association between predicted and true accuracies (range \([-1,1]\); higher is better).\\
\textbf{RMSE:} Root Mean Squared Error between predicted and true accuracies (percentage scale here); lower indicates better absolute agreement.

 We randomly select 30\% of the datasets as test sets to evaluate our Ridge Regression Model; however, we preselect African Food and Beans for the test set to test our method on underrepresented datasets as well. Table \ref{tab:clip_vs_prelabelprobe_error} reports the full-test-set CLIP zero-shot accuracy, our 1-shot \textit{PreLabellingProbe} estimates, and absolute errors. The estimates closely track the ground-truth accuracies (Pearson-r = 0.96; RMSE = 10.37), with low error on African Food and moderate underestimation on Beans and CIFAR-10.

\begin{table}[htbp]
\caption{Test Performance Comparison of Variants}
\begin{center}
\begin{tabular}{|c|c|c|}
\hline
\textbf{Variant} & \textbf{Pearson-r} & \textbf{RMSE} \\
\hline
LLM-generated only & 0.848922 & 0.144607 \\
\hline
Vanilla CLIP prompt-based only & 0.947374 & 0.150432 \\
\hline
PreLabellingProbe & \textbf{0.961896} & \textbf{0.103692} \\
\hline
\end{tabular}
\label{tab:performance_comparison}
\end{center}
\end{table}

\begin{table}[htbp]
\caption{Full Test Set CLIP zero-shot Accuracy vs. PreLabellingProbe (Percentage Scale). Pearson-r = 0.96. Root Mean Squared Error = 10.37.}
\begin{center}
\begin{tabular}{|c|c|c|c|}
\hline
\textbf{Dataset} & \textbf{\makecell{CLIP ZS Accuracy \\ (Full testset)}} & \textbf{\makecell{PreLabellingProbe \\(1-shot)}} & \textbf{Error} \\
\hline
African Food & 38.24 & 41.22 & 2.98 \\
\hline
Beans & 39.84 & 26.12 & -13.72 \\
\hline
Caltech101 & 89.25 & 84.86 & -4.39 \\
\hline
CIFAR10 & 91.68 & 76.41 & -15.27 \\
\hline
Food101 & 83.76 & 74.38 & -9.38 \\
\hline
\end{tabular}
\label{tab:clip_vs_prelabelprobe_error}
\end{center}
\end{table}

\section{Conclusion}
We introduce a novel, data-efficient method to estimate the dataset-level zero-shot performance of VLFMs. Using a single labelled example per class, we generate counterfactual captions using an LLM to probe how well the VLFM's latent representations are structured. Experiments show accurate performance estimation across diverse datasets, including specialised and underrepresented domains. This provides a practical, low-cost method to reduce unnecessary labelling.

\section{Acknowledgement}
This publication has emanated from research conducted with the financial support of Taighde Éireann – Research Ireland under Grant number 18/CRT/6183. For the purpose of Open Access, the author has applied a CC BY public copyright licence to any Author Accepted Manuscript version arising from this submission.

\bibliographystyle{ieeetr}
\bibliography{references}

\end{document}